\documentclass[11pt]{article}
\usepackage[hyperref]{ccl2025-en}
\usepackage{times}
\usepackage{url}
\usepackage{latexsym}
\usepackage{fancyhdr}
\usepackage{graphicx}
\usepackage{amsmath}
\usepackage{CJKutf8}
\usepackage{booktabs}
\usepackage{hyperref}

\title{System Report for CCL25-Eval Task 11: Aesthetic Assessment of Chinese Handwritings Based on Vision Language Models}

\author{
\begin{tabular}{@{}c}
Chen Zheng\textsuperscript{1,2}, ~~Yuxuan Lai\textsuperscript{1,2}, ~~Haoyang Lu\textsuperscript{3}, ~~Wentao Ma\textsuperscript{3}, \\
~~Jitao Yang\textsuperscript{3}, and Jian Wang\textsuperscript{3} \\
\normalfont\textsuperscript{1}The Open University of China, Beijing, China \\
\normalfont\textsuperscript{2}Engineering Research Center of Integration and Application of Digital Learning Technology, \\
\normalfont Ministry of Education, Beijing, China \\
\normalfont\textsuperscript{3}OUC-online, Beijing, China \\
{\tt zhengchen@ouchn.edu.cn}
\end{tabular}
}

\date{}

\begin{document}
\maketitle
\begin{abstract}
The handwriting of Chinese characters is a fundamental aspect of learning the Chinese language. Previous automated assessment methods often framed scoring as a regression problem. However, this score-only feedback lacks actionable guidance, which limits its effectiveness in helping learners improve their handwriting skills. In this paper, we leverage vision-language models (VLMs) to analyze the quality of handwritten Chinese characters and generate multi-level feedback. Specifically, we investigate two feedback generation tasks: simple grade feedback (Task 1) and enriched, descriptive feedback (Task 2). We explore both low-rank adaptation (LoRA)-based fine-tuning strategies and in-context learning methods to integrate aesthetic assessment knowledge into VLMs. Experimental results show that our approach achieves state-of-the-art performances across multiple evaluation tracks in the CCL 2025 workshop on evaluation of handwritten Chinese character quality.
  \englishkeywords{Handwritten Chinese Characters \and Aesthetic Assessment \and Vision-Language Models \and Low-rank Adaptation \and In-context Learning}
\end{abstract}

\section{Introduction}
\label{intro}

%
% The following footnote without marker is needed for the camera-ready
% version of the paper.
% Comment out the instructions (first text) and uncomment the 8 lines
% under "final paper" for your variant of English.
%
\cclfootnote{
    %
    % for review submission
    %
    \hspace{-0.65cm}  % space normally used by the marker
    % Place licence statement here for the camera-ready version. See Section~\ref{licence} of the instructions for preparing a manuscript.
    \textcopyright 2025 China National Conference on Computational Linguistics

    \noindent Published under Creative Commons Attribution 4.0 International License
}

The automated assessment of Chinese handwriting is a critical research area in language education and intelligent evaluation systems~\cite{XiaoLi:22,Chen:24}. Chinese handwritten characters, characterized by their linguistic accuracy and structural complexity, serve as a cornerstone of cultural and educational expression. However, existing systems typically provide only score-based feedback~\cite{Han:08,Gao:11,Li:14,Sun:15,Wang:16,Zhou:17,Wang:21,Sun:23,Wang:23,Yan:24,Wu:24}, which limits their effectiveness in supporting learners’ skill development. This highlights the need for advanced methods to deliver detailed, constructive feedback, thereby enhancing educational practices and supporting Chinese handwriting in digital learning environments.

Recent advancements in computer vision have facilitated the development of automated systems for evaluating Chinese handwriting, enabling standardized assessments while preserving the artistic qualities of calligraphy. However, constructing evaluation models that effectively balance standardization with aesthetic merit remains a complex challenge. Existing research has predominantly relied on hand-crafted features to assess structural and aesthetic quality. For instance, Gao et al. \shortcite{Gao:11} proposes a method for evaluating Chinese handwriting quality based on the recognition confidence of online handwriting analysis using a modified quadratic discriminant function classifier. Sun et al. \shortcite{Sun:15} utilize global shape features and component layout information to enhance aesthetic evaluation. Zhou et al. \shortcite{Zhou:17} use a possibility-probability distribution method to assess the quality of robotic Chinese handwriting. Despite these advances, such approaches often lack the flexibility to provide nuanced, context-aware feedback that effectively integrates both structural and stroke dimensions.

Recently, vision-language models (VLMs) have shown remarkable capabilities across various domains, including document understanding, visual perception, and multimodal reasoning~\cite{Qwen-VL,Qwen2-VL,Qwen2.5-VL,lu2024,kimivl}. Despite these advancements, their application in the aesthetic assessment of Chinese handwriting remains largely unexplored.

Traditional computer vision methods often struggle to provide fine-grained and personalized feedback in aesthetic assessment tasks. VLMs, with their robust capabilities in image understanding and natural language generation, offer a novel approach to address these limitations.

This study explores the application of VLMs to generate detailed, context-sensitive feedback on Chinese handwriting quality, with a focus on both structural integrity and stroke aesthetics. To effectively integrate domain-specific knowledge into VLMs for this task, we investigate two data-efficient methods: Low-Rank Adaptation (LoRA) based fine-tuning for open-source VLMs \cite{Hu:22}, and in-context learning for closed-source large language models (LLMs) \cite{Brown:20}.

We conducted experiments on the CCL 2025 evaluation task for assessing the quality of handwritten Chinese characters, which includes two subtasks: grading and comment generation. Our proposed method obtained scores of 0.76 and 0.52 on the respective subtasks, securing third place in the competition and demonstrating its effectiveness. 

\section{Task Formulation}

In the CCL 2025 evaluation of the quality of handwritten Chinese characters task, the objective is to assess the aesthetic quality of a given Chinese handwritten image. This task involves two distinct sub-tasks:

Task 1: Grading of handwritten Chinese characters: the goal is to classify the quality of handwritten characters into three discrete grades: excellent, medium, and unqualified. This classification is primarily based on the structural integrity and stroke aesthetics of the characters.

Task 2: Comment generation of handwritten Chinese characters: the goal is to provide targeted textual descriptions focusing on the two aforementioned dimensions: structure and stroke form.

\section{Methods}
We explore LoRA and in-context learning methods, with the overall framework depicted in Fig. 1.
\begin{figure}[ht!]
    \centering
    \includegraphics[width=0.95\textwidth]{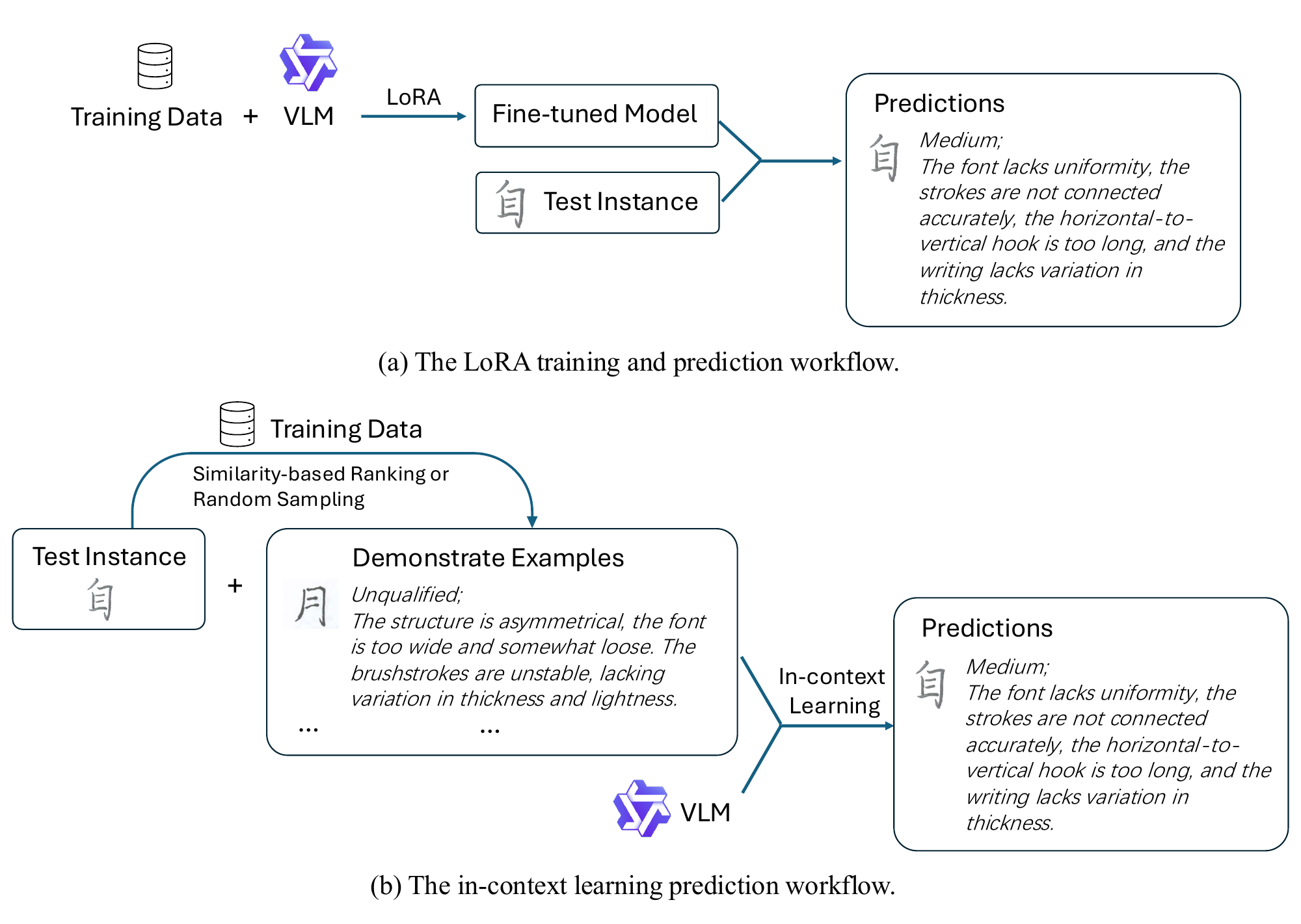} 
    \label{fig:p1} 
\caption{The LoRA and in-context learning frameworks.}
\end{figure}

\subsection{Training Format for LoRA}

For LoRA fine-tuning, training and testing data are structured as single-turn dialogues, following the template provided below. In Task 1, the model receives a raw image of a handwritten Chinese character as input, and its training objective is to output the corresponding quality grade.

\begin{quotation}
\noindent [ \{``role": ``user", ``content": \texttt{<INPUT\_IMAGE>}\}, \\
\{``role": ``assistant", ``content": \texttt{<GRADE>}\} ]
\end{quotation}

\noindent Task 2 employs two different input-output formats. The first is similar to Task 1, but the expected output is detailed feedback on handwriting quality, rather than just a grade. 

\begin{quotation}
\noindent [ \{``role": ``user", ``content": \texttt{<INPUT\_IMAGE>}\}, \\
\{``role": ``assistant", ``content": \texttt{<FEEDBACK>}\} ]
\end{quotation}

\noindent The second format's input differs by including the raw image of the handwritten Chinese character and the grade predicted by the model trained on Task 1.

\begin{quotation}
\noindent [ \{``role": ``user", ``content": \texttt{<INPUT\_IMAGE>} The evaluation for the above handwritten Chinese characters is \texttt{<GRADE>}, generate a comment.\}, \\
\{``role": ``assistant", ``content": \texttt{<FEEDBACK>}\} ]
\end{quotation}

\subsection{Example Demonstation for In-context Learning}

We investigated two in-context learning methods: a similarity-based method for selecting and ordering in-context examples, and random selection of in-context examples. In the first method, given a test instance, we select the $k$ most similar instances from the training data to serve as demonstrations. A training instance is placed closer to the test instance as its similarity increases. In the second method, instances are randomly selected from the training data.

The organization of the query for in-context learning is illustrated below:
\begin{quotation}
\noindent [ \{``role": ``system", ``content": \texttt{SYSTEM\_PROMPT}\}, \\
\{``role": ``user", ``content": \texttt{<INPUT\_IMAGE$_1$>}\}, \\
\{``role": ``assistant", ``content": \texttt{<GRADE$_1$>} or \texttt{<FEEDBACK$_1$>}\}, \\
..., \\
\{``role": ``user", ``content": \texttt{<INPUT\_IMAGE$_k$>}\}, \\
\{``role": ``assistant", ``content": \texttt{<GRADE$_k$>} or \texttt{<FEEDBACK$_k$>}\}, \\
\{``role": ``user", ``content": \texttt{TEST\_PROMPT}, \texttt{<TEST\_IMAGE>}\} ]
\end{quotation}
\noindent In the similarity-based method, the input images are ordered based on their similarity to a test image, as follows:
\begin{align}
{sim}(\texttt{INPUT\_IMAGE}_1, \texttt{TEST\_IMAGE}) &\leq {sim}(\texttt{INPUT\_IMAGE}_2, \texttt{TEST\_IMAGE}) \leq \ldots \nonumber \\
&\leq {sim}(\texttt{INPUT\_IMAGE}_k, \texttt{TEST\_IMAGE})
\end{align}
\noindent Here, $sim(\dot)$ denotes the cosine similarity between the image embeddings of the instances. The system prompt for Task 1 clearly outlined the evaluation criteria for each grade, as shown below \footnote{All prompts were originally in Chinese and have been translated into English for presentation.}. 
\begin{quotation}
\noindent \texttt{SYSTEM\_PROMPT}: You are an expert in Chinese calligraphy who is familiar with the aesthetic features of Chinese characters. You are capable of accurately evaluating the quality of students' handwriting. \\
Following the example provided, you are required to rate the given samples of Chinese character writing into three grades: A (Excellent), B (Medium), or C (Unqualified). The grading criteria are defined as follows:\\
A (Excellent): The character structure and proportions are well-balanced, the center of gravity is stable, and the overall appearance is symmetrical and aesthetically pleasing. The strokes are correctly shaped with proper coordination, clearly executed, and demonstrate variation in pressure and thickness. There are no significant flaws in the writing.\\
B (Medium): The structure and proportions are generally reasonable, the center of gravity is mostly stable, and the character appears relatively balanced. The stroke forms are largely correct, and the individual strokes are clear, but there is limited variation in pressure. The writing exhibits systematic deficiencies in one or more aspects.\\
C (Unqualified): The structure is imbalanced or the center of gravity is unstable, resulting in an asymmetrical and unappealing appearance. The strokes are careless, unclear, or sloppy. The writing contains serious flaws that significantly affect legibility or aesthetic quality.
\end{quotation}

\noindent The \texttt{SYSTEM\_PROMPT} for Task 2 outlined the key points for generating feedback, details of which can be found in Appendix A. The \texttt{TEST\_PROMPT} instructs the VLM to grade the test image or give feedback based on the instructions and examples. For Task 1,
\begin{quotation}
\noindent \texttt{TEST\_PROMPT}: You are required to assign a score of A (Excellent), B (Medium), or C (Unqualified) to the given image of Chinese character writing, based on the example and criteria provided above. Note: Your response must consist of only a single uppercase letter corresponding to the score for this image.
\end{quotation}
\noindent The details of the \texttt{TEST\_PROMPT} for task 2 can be found in appendix B.

\section{Experiment}

\subsection{Experimental Setups}

We conduct experiments on the CCL 2025 Evaluation of the quality of handwritten Chinese characters dataset. For Task 1, the dataset comprises 1500 training instances and 300 test instances. For Task 2, it includes 600 training instances and 100 test instances.

For Task 1, evaluation metrics include precision, recall and the F1-score. For Task 2, the metrics are ROUGE-1, ROUGE-2, and ROUGE-L. The final score is calculated as follows: 

\begin{equation}
\text{FinalScore} = 0.4 \times \text{ROUGE-L} + 0.3 \times \text{ROUGE-2} + 0.3 \times \text{ROUGE-1}.
\end{equation}

We use the open-source VLMs \textit{Qwen2.5-VL-72B-Instruct}\footnote{\url{https://huggingface.co/Qwen/Qwen2.5-VL-72B-Instruct}} (QwenVL) (for both task 1 and 2) and \textit{QVQ-72B-Preview}\footnote{\url{https://huggingface.co/Qwen/QVQ-72B-Preview}} (QVQ) (for task 2) as the base models for LoRA training. The training was conducted for 3 epochs with a learning rate of $1 \times 10^{-4}$ using the open-source fine-tuning tool LLaMA-Factory \cite{zheng2024}. 

In the LoRA training for Task 1, we utilized an open-source dataset, CHAED \cite{Sun:15} to expand our training set. This dataset comprises 1000 Chinese handwriting images, each accompanied by aesthetic scores. Images were empirically classified into three categories based on their aesthetic scores: \textit{Excellent} (scores $>$ 80), \textit{Medium} (scores 30 $\sim$ 80), and \textit{Unqualified} (scores $<$ 30). Separate models were then trained using only the task-specific dataset and with the combined CHAED data, respectively.

For LoRA-based training in Task 2, the first model was trained similarly to Task 1, but it generated feedback text instead of grading scores. The second model utilized the model trained on the Task 1-specific dataset to predict grading scores for each image in the training and test sets. Subsequently, the handwriting images with their predicted grades were used as input, while the corresponding feedback text served as the output to fine-tune the model.

In the in-context learning strategy, we compared the performance of similarity-based ordering for in-context examples against random selection of in-context examples. The model used in the in-context learning is the closed-source VLM \textit{qwen-vl-max-2025-01-25}\footnote{\url{https://bailian.console.aliyun.com/?tab=model\#/model-market/detail/qwen-vl-max?modelGroup=qwen-vl-max}}. For the selection and ordering of in-context examples, we use the \textit{multimodal-embedding-v1}\footnote{\url{https://bailian.console.aliyun.com/?tab=model\#/model-market/detail/multimodal-embedding-v1}} provided by Alibaba Cloud for image embedding. Vector indexing was implemented with ChromaDB\footnote{\url{https://www.trychroma.com}}. 

For Task 1, we separated 300 examples from the training set as a development set and found that similarity-based ordering of in-context examples performed better. In Task 2, we separated 100 examples from the training set as a development set and found that random selection of in-context examples performed better.

\subsection{Results}

Table 1 and 2 presents the main results. The results for Task 1 indicate that the model fine-tuned on the task-specific dataset achieved the best performance. However, the model fine-tuned on the expanded dataset exhibited suboptimal performance, likely because the aesthetic score classification was misaligned with the grading criteria of the task-specific dataset.

The results for Task 2 demonstrate that the in-context learning method achieved the best performance. The fine-tuned QVQ model outperformed the QwenVL model. Additionally, the model trained with images paired with their predicted grades showed a marginal improvement of 0.03 in final score.

\begin{table}[t!]
\centering
\setlength\tabcolsep{10pt}
\label{tab:mainresults1}
\begin{tabular}{lccc}
\toprule
Model & Precision & Recall & F1 \\ \midrule
QwenVL LoRA & \textbf{0.76} & \textbf{0.76} & \textbf{0.76} \\ 
QwenVL LoRA w/ CHAED & 0.61 & 0.61 & 0.61 \\ 
In-Context Learning & 0.69 & 0.69 & 0.69 \\ 
\bottomrule
\end{tabular}
\caption{Summary of results of the task 1.}
\end{table}

\begin{table}[t!]
\centering
\setlength\tabcolsep{10pt}
% \small
\label{tab:mainresults2}
\begin{tabular}{lcccc}
\toprule
Model & ROUGE-1 & ROUGE-2 & ROUGE-L & FinalScore \\ \midrule
QwenVL LoRA & 0.43 & 0.24 & 0.41 & 0.36 \\ 
QwenVL LoRA w/ grade & 0.46 & 0.26 & 0.43 & 0.39 \\ 
QVQ LoRA  & 0.47 & 0.26 & 0.44 & 0.39 \\ 
In-Context Learning & \textbf{0.63} & \textbf{0.34} & \textbf{0.56} & \textbf{0.52} \\ 
\bottomrule
\end{tabular}
\caption{Summary of results of the task 2.}
\end{table}

\section{Conclusion and Future Work}
In this paper, we explore the application of VLMs to the evaluation of Chinese handwritten characters. Utilizing both open-source and closed-source VLMs, we investigate multiple strategies, including LoRA and in-context learning. Our approach achieved third place on the final leaderboard, demonstrating the effectiveness of the proposed methods.

In practical applications, fine-tuning VLMs is more computationally efficient than in-context learning, as the latter requires significantly higher token consumption and computational resources.

Building on recent advancements in reinforcement learning (RL) for training LLMs and VLMs \cite{Guo:25,Team:25}, our future work will focus on advancing the aesthetic assessment capabilities of VLMs through two directions. First, we will design comparative ranking tasks and fine-grained classification tasks to enhance the precision of aesthetic assessments in handwritten Chinese characters. Second, we will explore RL's potential in reasoning about complex aesthetic principles, while tackling challenges related to subjective evaluation and data scarcity.

\section*{Acknowledgements}

This work is supported by the National Natural Science Foundation of China (NSFC) under Grant Nos. 82371397 and 62206070 and the Innovation Fund Project of the Engineering Research Center of Integration and Application of Digital Learning Technology, Ministry of Education  via Grant No.1421012. We thank the Open University of China for the computational resources provided by its AI infrastructure.

%\bibliographystyle{ccl}
%\bibliography{ccl2025-en}

\begin{thebibliography}{}
\bibitem[\protect\citename{Sun et al.}2015]{Sun:15}
Rongju Sun, Zhouhui Lian, Yingmin Tang, and Jianguo Xiao.
\newblock 2015.
\newblock Aesthetic Visual Quality Evaluation of Chinese Handwritings.
\newblock In {\em Proceedings of the International Joint Conference on Artificial Intelligence (IJCAI)}, volume~15, pages 2510--2516.
  
\bibitem[\protect\citename{Xiao et al.}2022]{XiaoLi:22}
Xue Xiao and Chengcheng Li.
\newblock 2022.
\newblock Research Progress on Evaluation Methods of Handwritten Chinese Characters.
\newblock {\em Computer Engineering and Applications}, 58(2):27-42.

\bibitem[\protect\citename{Chen et~al.}2024]{Chen:24}
Weiran Chen, Jiaqi Su, Weitao Song, Jialiang Xu, Guiqian Zhu, Ying Li, Yi Ji, and Chunping Liu.
\newblock 2024.
\newblock Quality Evaluation Methods of Handwritten Chinese Characters: A Comprehensive Survey.
\newblock {\em Multimedia Systems}, 30(4):194.

\bibitem[\protect\citename{Yan et al.}2024]{Yan:24}
Fei Yan, Xueping Lan, Hua Zhang, and Linjing Li.
\newblock 2024.
\newblock Intelligent Evaluation of Chinese Hard-Pen Calligraphy Using a Siamese Transformer Network.
\newblock {\em Applied Sciences} 14, no. 5: 2051.

\bibitem[\protect\citename{Han et al.}2008]{Han:08}
Chin-Chuan Han, Chih-Hsun Chou, and Chung-Shiou Wu.
\newblock 2008.
\newblock An Interactive Grading and Learning System for Chinese Calligraphy.
\newblock {\em Machine Vision and Applications} 19:43--55.

\bibitem[\protect\citename{Gao et al.}2011]{Gao:11}
Yan~Gao, Lianwen~Jin, and Nanxi~Li.
\newblock 2011.
\newblock Chinese Handwriting Quality Evaluation Based on Analysis of Recognition Confidence.
\newblock In {\em 2011 IEEE International Conference on Information and Automation}, 221--225.
\newblock IEEE.

\bibitem[\protect\citename{Li et al.}2014]{Li:14}
Wei Li, Yuping Song, and Changle Zhou.
\newblock 2014.
\newblock Computationally Evaluating and Synthesizing Chinese Calligraphy.
\newblock {\em Neurocomputing} 135: 299--305.


\bibitem[\protect\citename{Wang et al.}2016]{Wang:16}
Mengdi Wang, Qian Fu, Xingce Wang, Zhongke Wu, and Mingquan Zhou.
\newblock 2016.
\newblock Evaluation of Chinese Calligraphy by Using DBSC Vectorization and ICP Algorithm.
\newblock {\em Mathematical Problems in Engineering} 2016, 1:4845092.

\bibitem[\protect\citename{Zhou et al.}2017]{Zhou:17}
Dajun Zhou, Jiamin Ge, Ruiqi Wu, Fei Chao, Longzhi Yang, and Changle Zhou.
\newblock 2017.
\newblock A Computational Evaluation System of Chinese Calligraphy via Extended Possibility-Probability Distribution Method.
\newblock In {\em 2017 13th International Conference on Natural Computation, Fuzzy Systems and Knowledge Discovery (ICNC-FSKD)}, pages 884--889.
\newblock IEEE.

\bibitem[\protect\citename{Sun et al.}2023]{Sun:23}
Mingwei Sun, Xinyu Gong, Haitao Nie, Muhammad Minhas Iqbal, and Bin Xie.
\newblock 2023.
\newblock SRAFE: Siamese Regression Aesthetic Fusion Evaluation for Chinese Calligraphic Copy.
\newblock {\em CAAI Transactions on Intelligence Technology} 8, no. 3: 1077-1086.

\bibitem[\protect\citename{Wang and Lv}2021]{Wang:21}
Zhaoyi Wang and Ruimin Lv.
\newblock 2021.
\newblock Design of Calligraphy Aesthetic Evaluation Model Based on Deep Learning and Writing Action.
\newblock In {\em International Conference on Computing, Control and Industrial Engineering}, pp. 620--628. Singapore: Springer Nature Singapore.

\bibitem[\protect\citename{Wang et~al.}2023]{Wang:23}
Min Wang, Wan Ma, Chuang Zhu, Shanfei Shi, Jiangbo Shu, and Shuaicheng Lu.
\newblock 2023.
\newblock Research on Quantitative Evaluation of Standard Chinese Characters Written by Pen and Paper Based on Neural Network.
\newblock {\em Journal of Central China Normal University (Natural Sciences)}, 57(6): 813--820.

\bibitem[\protect\citename{Wu et~al.}2024]{Wu:24}
Meng-Luen Wu, Yi-Rong Du, and Dai-Hua Jiang.
\newblock 2024.
\newblock Aesthetic Evaluation System for Calligraphy Characters using Convolutional Neural Networks.
\newblock In {\em 2024 International Conference on Machine Learning and Cybernetics (ICMLC)}, pp. 547--552. IEEE.

\bibitem[\protect\citename{Bai et~al.}2023]{Qwen-VL}
Jinze Bai, Shuai Bai, Shusheng Yang, Shijie Wang, Sinan Tan, Peng Wang, Junyang Lin, Chang Zhou and Jingren Zhou.
\newblock 2023.
\newblock Qwen-VL: A Versatile Vision-Language Model for Understanding, Localization, Text Reading, and Beyond.
\newblock {\em arXiv:2308.12966}.

\bibitem[\protect\citename{Wang et~al.}2024]{Qwen2-VL}
Peng Wang, Shuai Bai, Sinan Tan, Shijie Wang, Zhihao Fan, Jinze Bai, Keqin Chen, Xuejing Liu, Jialin Wang, Wenbin Ge, Yang Fan, Kai Dang, Mengfei Du, Xuancheng Ren, Rui Men, Dayiheng Liu, Chang Zhou, Jingren Zhou, and Junyang Lin.
\newblock 2024.
\newblock Qwen2-VL: Enhancing Vision-Language Model's Perception of the World at Any Resolution.
\newblock {\em arXiv:2409.12191}.

\bibitem[\protect\citename{Bai et~al.}2025]{Qwen2.5-VL}
Shuai Bai, Keqin Chen, Xuejing Liu, Jialin Wang, Wenbin Ge, Sibo Song, Kai Dang, Peng Wang, Shijie Wang, Jun Tang, Humen Zhong, Yuanzhi Zhu, Mingkun Yang, Zhaohai Li, Jianqiang Wan, Pengfei Wang, Wei Ding, Zheren Fu, Yiheng Xu, Jiabo Ye, Xi Zhang, Tianbao Xie, Zesen Cheng, Hang Zhang, Zhibo Yang, Haiyang Xu, and Junyang Lin.
\newblock 2025.
\newblock Qwen2.5-VL Technical Report.
\newblock {\em arXiv:2502.13923}.

\bibitem[\protect\citename{Kimi Team et~al.}2025]{kimivl}
Kimi Team, Angang Du, Bohong Yin, et~al.
\newblock 2025.
\newblock Kimi-VL Technical Report.
\newblock {\em arXiv:2504.07491}.

\bibitem[\protect\citename{Lu et~al.}2024]{lu2024}
Haoyu Lu, Wen Liu, Bo Zhang, Bingxuan Wang, Kai Dong, Bo Liu, Jingxiang Sun, Tongzheng Ren, Zhuoshu Li, Hao Yang, Yaofeng Sun, Chengqi Deng, Hanwei Xu, Zhenda Xie, and Chong Ruan.
\newblock 2024.
\newblock DeepSeek-VL: Towards Real-World Vision-Language Understanding.
\newblock {\em arXiv:2403.05525}.

\bibitem[\protect\citename{Hu et~al.}2022]{Hu:22}
Edward~J. Hu, Yelong Shen, Phillip Wallis, Zeyuan Allen-Zhu, Yuanzhi Li, Shean Wang, Lu Wang, and Weizhu Chen.
\newblock 2022.
\newblock LoRA: Low-Rank Adaptation of Large Language Models.
\newblock {\em ICLR} 1(2): 3.

\bibitem[\protect\citename{Brown et~al.}2020]{Brown:20}
Tom Brown, Benjamin Mann, Nick Ryder, et~al.
\newblock 2020.
\newblock Language Models are Few-Shot Learners.
\newblock {\em Advances in Neural Information Processing Systems}, 33: 1877--1901.

\bibitem[\protect\citename{Zheng et~al.}2024]{zheng2024}
Yaowei~Zheng, Richong~Zhang, Junhao~Zhang, Yanhan~Ye, Zheyan~Luo, Zhangchi~Feng, and Yongqiang~Ma.
\newblock 2024.
\newblock LlamaFactory: Unified Efficient Fine-Tuning of 100+ Language Models.
\newblock In {\em Proceedings of the 62nd Annual Meeting of the Association for Computational Linguistics (Volume 3: System Demonstrations)}, Bangkok, Thailand. Association for Computational Linguistics.

\bibitem[\protect\citename{Guo et al.}2025]{Guo:25}
DeepSeek-AI, Daya Guo, Dejian Yang, et al.
\newblock 2025.
\newblock DeepSeek-R1: Incentivizing Reasoning Capability in LLMs via Reinforcement Learning.
\newblock {\em arXiv:2501.12948}.

\bibitem[\protect\citename{Kimi Team et al.}2025]{Team:25}
Kimi Team, Angang Du, Bofei Gao, et al.
\newblock 2025.
\newblock Kimi k1.5: Scaling Reinforcement Learning with LLMs.
\newblock {\em arXiv:2501.12599}.

\end{thebibliography}

\appendix
\begin{CJK*}{UTF8}{gbsn}
\section{System prompts}

For Task 1, the original \texttt{SYSTEM\_PROMPT} in the in-context learning method was written in Chinese: 

\begin{quotation}
你是一名汉字书法专家，你对汉字的图形图像非常了解，可以准确评价学生汉字书写的质量。
你需要仿照样例，对给出的汉字书写图片按A: 优秀，B: 中等，C: 不合格 三个等级打分。三个等级的评价标准如下：\\
A. 优秀：结构比例安排适当，重心平稳，字体匀称美观。点画形态正确且有呼应，笔画清晰到位，用笔有轻重变化之分。字体无明显缺陷。\\
B. 中等：结构比例基本合理，重心基本稳定，字体较匀称。笔画形态基本正确，点画清晰但轻重变化不够。字体在某一类或几类问题上存在系统性缺陷。\\
C. 不合格：结构比例失衡或重心不稳，字体不匀称。点画随意，笔画不清晰或潦草。字体存在较严重缺点。
\end{quotation}

\noindent The English version used in experiments is provided in Section 3.2. For Task 2, the \texttt{SYSTEM\_PROMPT} was: 

\begin{quotation}
You are an expert in Chinese calligraphy with a deep understanding of the graphical aspects of Chinese characters, capable of accurately evaluating the quality of students' handwriting. You need to follow the examples and write comments for the given Chinese character images. The comments should provide targeted evaluations and descriptions focusing on two main dimensions: structure and stroke form.\\
For structure, consider density, balance (such as the symmetry of top-bottom or left-right structures), and the center of gravity.\\
For strokes, consider the variation in stroke weight and the specific forms of individual strokes.
\end{quotation}

\noindent The original Chinese version: 

\begin{quotation}
你是一名汉字书法专家，你对汉字的图形图像非常了解，可以准确评价学生汉字书写的质量。
你需要仿照样例，对给定的汉字图片撰写评语。评语主要对结构和笔画形态两大维度，进行有针对性的评价和描述。\\
结构上，考虑疏密、匀称（如上下结构、左右结构等方面的匀称性）、重心。\\
笔画上，考虑笔画的轻重变化，以及具体笔画的形态。
\end{quotation}

\section{Test prompts}

The \texttt{TEST\_PROMPT} for task 1 in the in-context learning method was:

\begin{quotation}
You need to refer to the above image and scoring to grade the Chinese character writing in the image below as A: Excellent, B: Medium, C: Unqualified. \\
Attention! Your output must only contain one uppercase letter! Corresponding to the score of the Chinese character writing in this image.
\end{quotation}

\noindent The original Chinese version: 

\begin{quotation}
你需要参照上面的图片及打分，对下面这张汉字书写的图片按照A: 优秀，B: 中等，C: 不合格  给出分数。\\
注意！你的输出只能有一个大写字母！对应这张图上汉字书写的分数。
\end{quotation}

\noindent The \texttt{TEST\_PROMPT} for task 2 was:

\begin{quotation}
You need to refer to the above image and the corresponding comments to write a critique for the following Chinese handwriting image. \\
Attention! Your output format and content style must strictly follow the reference comments. Write a passage of similar length and style. 
\end{quotation}

\noindent The original Chinese version: 

\begin{quotation}
你需要参照上面的图片及对应的评语，对下面这张汉字书写的图片撰写评语。\\
注意！你输出的格式和内容风格要严格参考上面的评语。以相似的长度和风格撰写一段话。
\end{quotation}
\end{CJK*}

\end{document}